\documentclass[letterpaper]{article} 
\usepackage{aaai23}  
\usepackage{times}  
\usepackage{helvet}  
\usepackage{courier}  
\usepackage[hyphens]{url}  
\usepackage{graphicx} 
\usepackage{natbib}  
\usepackage{caption} 
\usepackage{algorithm}
\usepackage{algorithmic}
\usepackage{caption}
\usepackage{subcaption}
\usepackage{newfloat}
\usepackage{listings}
\DeclareCaptionStyle{ruled}{labelfont=normalfont,labelsep=colon,strut=off} 
\floatstyle{ruled}
\newfloat{listing}{tb}{lst}{}
\floatname{listing}{Listing}
\title{Explaining Large Language Model-Based Neural Semantic Parsers (Student Abstract)}

\author{
    Daking Rai,\textsuperscript{\rm 1}
    Yilun Zhou,\textsuperscript{\rm 2}
    Bailin Wang,\textsuperscript{\rm 2}
    Ziyu Yao\textsuperscript{\rm 1}
}
\affiliations{
    \textsuperscript{\rm 1} George Mason University, \\
    \textsuperscript{\rm 2} Massachusetts Institute of Technology\\
     drai2@gmu.edu,	yilun@mit.edu, bailinw@mit.edu, ziyuyao@gmu.edu
}

\usepackage{bibentry}

\begin{document}

\maketitle

\begin{abstract}
While large language models (LLMs) have demonstrated strong capability in structured prediction tasks such as semantic parsing, few amounts of research have explored the underlying mechanisms of their success. Our work studies different methods for explaining an LLM-based semantic parser and qualitatively discusses the explained model behaviors, hoping to inspire future research toward better understanding them.
\end{abstract}

\section{Introduction}

Semantic parsing is a task of mapping natural language utterances to their logical forms like SQL queries or lambda expressions for database or knowledge base querying. Despite its structured prediction nature, recent work has shown that a large language model (LLM) which generates output sequentially could achieve comparable or even better performance than the traditional structured decoders \cite{Scholak2021:PICARD}. However, why these LLMs could do well in semantic parsing is still unclear. 

In this paper, we seek to provide one of the first studies toward explaining LLM-based neural semantic parsers. We use the text-to-SQL semantic parsing task \cite{yu-etal-2018-spider} and the UnifiedSKG model \cite{UnifiedSKG} for a case study. We empirically {explore} a set of local explanation methods and quantitatively discussed the explanation results.

\section{Method}

(1) LIME \cite{10.1145/2939672.2939778} generates an explanation by training locally-faithful interpretable models with the dataset obtained by perturbing the prediction instance. 
(2) Shapley value measures the importance of a feature by its average marginal contribution to the prediction score. 
(3) Kernel SHAP \cite{NIPS2017_7062} is another efficient way of estimating Shapley values by training a linear classifier. (4) LERG \cite{tuan2021local} is a set of two approaches, LERG\_L and LERG\_S, recently adapted from LIME and Shapley value to conditioned sequence generation tasks. 
When applying these methods to explain an LLM-based semantic parser, we consider each output token as one prediction and attribute it to the input features.
(5) Attention: Prior work has revealed that attention may be interpreted as feature importance. Therefore, we also introduce an attention-based local explanation method, where the feature attribution is calculated by averaging the last layer of the multi-headed cross-attention weights. 

\subsection{Experimental Setup}

\begin{figure}[t!]
\centering
\includegraphics[width=1\columnwidth]{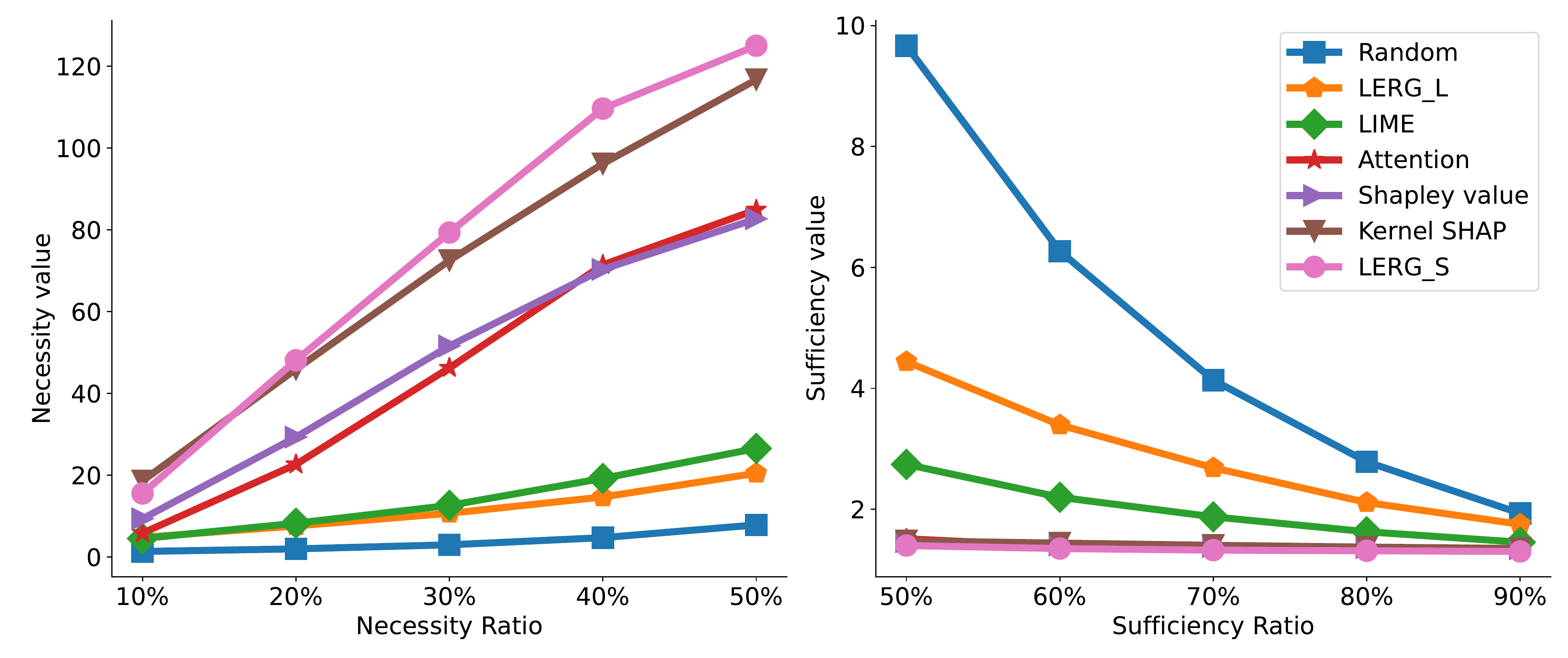} 
\caption{Necessity (left) and Sufficiency (right) scores when removing or keeping the top-K\% important features.
}
\label{fig1}
\end{figure}

\begin{figure*}[t!]
    \centering
    \subfloat[\centering Correct Prediction]
    {{\includegraphics[width=0.35\textwidth]   {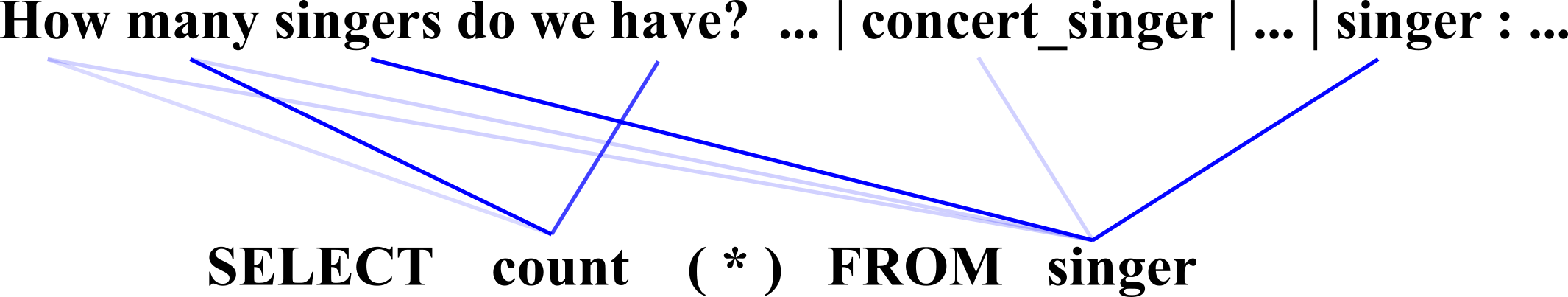} }}%
 \qquad
\subfloat[\centering Incorrect Prediction]
    {{\includegraphics[width=0.55\textwidth]{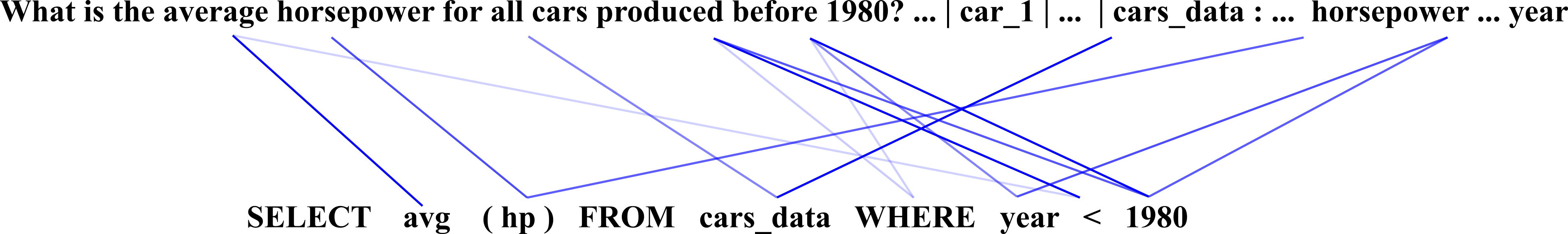} }}%
    \qquad
    \subfloat[\centering In-domain Example ]{{\includegraphics[width=0.3\textwidth]{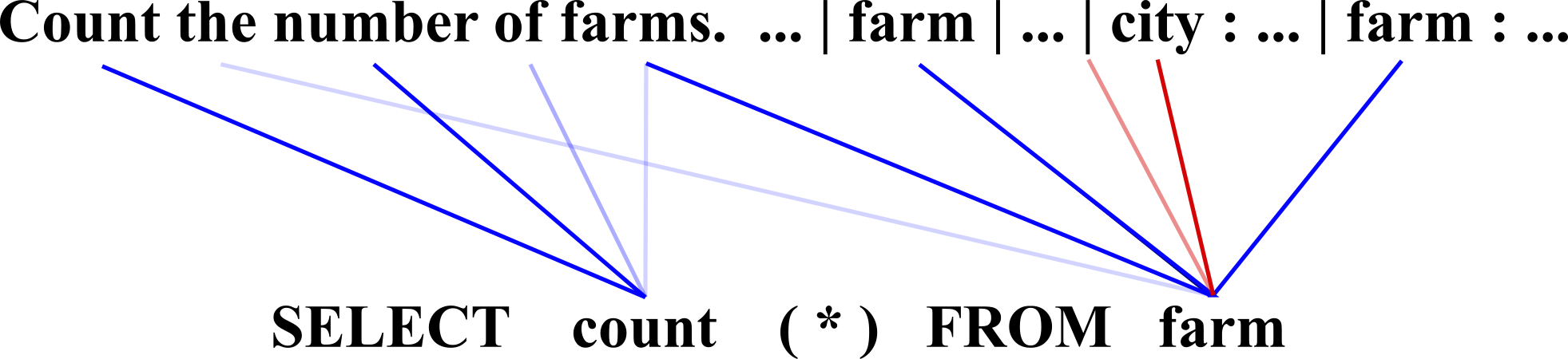} }}%
    \qquad
\subfloat[\centering Compositional Generalization/Hard-level Example]
    {{\includegraphics[width=0.6\textwidth]{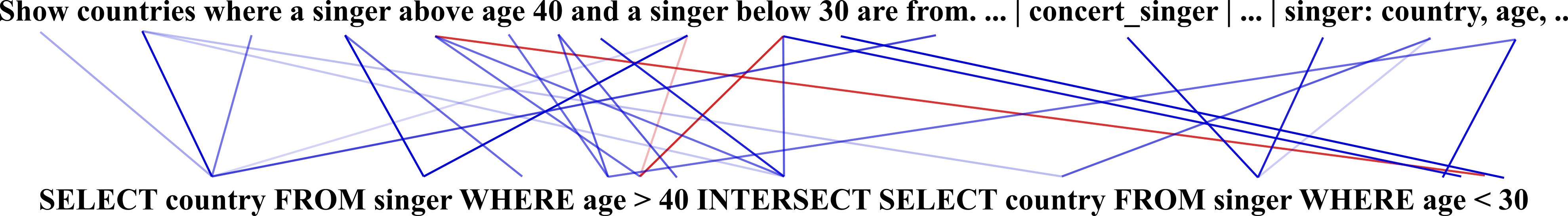} }}%

    \caption{
    UnifiedSKG generally shows plausible explanations. 
    In (a), ``concert\_singer'' is the database name; ``singer : ...'' shows the table along with its columns (omitted with ellipsis); similarly for other examples.
    For (b), the selected column ``hp'' is incorrect and should be ``horsepower''.
    Blue/red indicates positive/negative importance. Darkness indicates strength.
    }%
\label{fig2}
\end{figure*}

In our experiments, we consider the task of text-to-SQL semantic parsing where the goal is to generate a SQL query given a natural language question and the database schema (i.e., tables and columns included in the database) {as input}. We experiment with UnifiedSKG \cite{UnifiedSKG}, one of the state-of-the-art models, which adopts a T5 encoder-decoder structure.\footnote{We used the ``T5\_base\_prefix\_spider\_with\_cell\_value'' version from \url{https://github.com/HKUNLP/UnifiedSKG}. We did not use the T5-3B version because of the large computational demand, which we will discuss in Section~\ref{sec:discussion}.}
Following \citet{UnifiedSKG}, we train and evaluate the parser on the Spider dataset \cite{yu-etal-2018-spider}.

Through the experiments, we seek to answer two \emph{Research Questions (RQs)}: (1)\textit{  Which local explanation method is the most faithful to explaining the LLM-based UnifiedSKG parser?} (2) \textit{ How well does the explanation align with human intuitions?  }
To answer RQ1, we follow \citet{tuan2021local} and compare different explanation methods on two metrics:
(a) Sufficiency measures the perplexity when keeping only the top-K\% most important features by each explanation method; {the lower the better/faithful}. 
(b) Necessity measures the perplexity change when the top-K\% most important features are removed; {the higher the better/faithful}.
To answer RQ2, we qualitatively discuss the most faithful explanation results.

\subsection{Experimental Results}

\textbf{Faithfulness.} The results in Figure~\ref{fig1} show LERG\_S has the best performance as per both sufficiency and necessity metrics with Kernel SHAP having comparable performance as well. In general, we observe that Shapley value-based explanation methods have more faithful explanations than other methods. In addition, we also found that attention-based explanations are more faithful than LIME and LERG\_L.

\paragraph{Plausibility.} Using LERG\_S as a lens, we qualitatively study how UnifiedSKG works. We define \emph{plausible} explanations as those which align well with human intuition.
In our study, we classify each explanation into plausible or partially plausible ones. Interestingly, we didn't find any explanation that is completely implausible. Under this setup, we investigate the four aspects listed below (Figure~\ref{fig2}):
\textbf{(1) Feature Attribution for (In)correct Predictions}:
We randomly sample 20 examples where the model makes correct and incorrect predictions, respectively. {We find out that in most cases (85\% for correct and 70\% for incorrect), LERG\_S generates a plausible explanation for both types.} \textbf{(2) Different Hardness Levels}: We randomly sample 20 examples for each hardness level -- easy, medium, hard, and extra hard, as defined by the Spider benchmark based on the SQL complexity. We observed that in most cases the model behaviors are in line with human intuitions even at the extra hard level (80\%; $>$90\% for other levels). \textbf{(3) Compositional Generalization}: {We seek to understand whether the model attributes the output fragments to correct features compositionally when it makes correct predictions.} We conducted a similar manual examination as before and observed that in most (80\%) cases our model shows compositionally generalizable feature attribution. \textbf{(4) In-domain vs. Out-of-domain}: As the Spider training and dev sets are split by databases (which could be seen as different domains), we also manually compare the model explanations in in-domain and out-of-domain cases. We observe that for both cases (75\% and 80\% respectively), the generated explanations were mostly plausible.

\section{Discussion and Future Directions}\label{sec:discussion}

Our study has revealed several challenges and opportunities in explaining an LLM-based semantic parser: 
\textbf{(1) Computational costs}: Most local explanation methods require model inference over a large set of input perturbations, which is computationally inefficient. Future work may look into improving the attention-based explanation method, which does not rely on perturbations and hence could save much computation. \textbf{(2) Feature interaction}: Traditional feature attribution does not provide information about how features (e.g., question tokens and contextual database schema items) interact with each other. Future work may uncover these interactions to gain deeper insights into how the model works.
\textbf{(3) Explanation for user understanding}: Current saliency maps encompass a lot of information. Future work could examine how to present the information in a concise and friendly way such that users could easily grasp the intuition of the model prediction and verify its correctness. 
\textbf{(4) Explanation for debugging}: Future work should also investigate how the local explanation results could be used to probe and debug a semantic parser, such as to improve their capability in compositional generalization.


\fontsize{9.0pt}{10.0pt} \selectfont

\end{document}